# Interação entre Robôs Humanoides: Desenvolvendo a Colaboração e Comunicação Autônoma

Moraes Pablo, pablo.moraes@utec.edu.uy[1]
Rodríguez Mónica, monica.rodriguez@utec.edu.uy[1]
Peters Christopher, qristopherp@gmail.com[2]
Sodre Hiago, hiago.sodre@utec.edu.uy[1]
Mazondo Ahilen, ahilen.mazondo@estudiantes.utec.edu.uy[1]
Sandin Vincent, vincent.sandin@estudiantes.utec.edu.uy[1]
Barcelona Sebastian, sebastian.barcelona@utec.edu.uy[1]
Moraes William, william.moraes@estudiantes.utec.edu.uy[1]
Fernández Santiago, santiago.fernandez@estudiantes.utec.edu.uy[1]

Assunção Nathalie, nathalie.assuncao@utec.edu.uy[1]
de Vargas Bruna, bruna.devargas@utec.edu.uy[1]
Dörnbach Tobias, t.doernbach@ostfalia.de[2]
Kelbouscas André, andre.dasilva@utec.edu.uy[1]
Grando Ricardo, ricardo.bedin@utec.edu.uy[1]

[1]Universidad Tecnológica del Uruguay
[2]Ostfalia University of Applied Sciences

***Abstract:*** *This study investigates the interaction between humanoid robots NAO and Pepper, emphasizing their potential applications in educational settings. NAO, widely used in education, and Pepper, designed for social interactions, offer new opportunities for autonomous communication and collaboration. Through a series of programmed interactions, the robots demonstrated their ability to communicate and coordinate actions autonomously, highlighting their potential as tools for enhancing learning environments. The research also explores the integration of emerging technologies, such as artificial intelligence, into these systems, allowing robots to learn from each other and adapt their behavior. The findings suggest that NAO and Pepper can significantly contribute to both technical learning and the development of social and emotional skills in students, offering innovative pedagogical approaches through the use of humanoid robotics.* ***Demonstration video:*** *video*

***Resumo:*** Este estudo investiga a interação entre os robôs humanoides NAO e Pepper, com ênfase em suas potenciais aplicações em contextos educacionais. O NAO, amplamente utilizado em ambientes educativos, e o Pepper, projetado para interações sociais, oferecem novas oportunidades para a comunicação e colaboração autônoma. Através de uma série de interações programadas, os robôs demonstraram sua capacidade de se comunicar e coordenar ações de forma autônoma, destacando seu potencial como ferramentas para aprimorar os ambientes de aprendizagem. A pesquisa também explora a integração de tecnologias emergentes, como a inteligência artificial, nesses sistemas, permitindo que os robôs aprendam uns com os outros e adaptem seu comportamento. Os resultados sugerem que o NAO e o Pepper podem contribuir significativamente tanto para o aprendizado técnico quanto para o desenvolvimento de habilidades sociais e emocionais nos estudantes, oferecendo abordagens pedagógicas inovadoras por meio do uso da robótica humanoide. **Demonstração visual:** vídeo.

Palavras-chave: Robôs Humanoides, NAO, Pepper, Comunicação Autônoma, Robótica Educacional, Interação Social

## 1. INTRODUÇÃO

Nos últimos anos, a robótica humanoide tem avançado significativamente, tanto em termos de capacidades técnicas quanto na maneira como os robôs interagem entre si e com humanos. Entre essas inovações, destaca-se a interação entre robôs, como NAO, um robô compacto e versátil amplamente utilizado em contextos educativos, e Pepper, projetado para interações sociais em ambientes públicos. Juntos, esses robôs abrem novas possibilidades para a comunicação e a colaboração autônoma (Mubin, 2018). Este estudo busca explorar os primeiros passos nessa direção, focando na interação entre NAO e Pepper.

A interação entre robôs humanoides é crucial para o desenvolvimento de sistemas que funcionem de maneira colaborativa em ambientes reais. A capacidade desses robôs de se comunicar e coordenar suas ações é vista como um elemento fundamental para melhorar a eficiência em diversas aplicações, desde a educação até a automação industrial. Além disso, a interação entre robôs humanoides está na vanguarda de uma nova era na robótica, onde a cooperação entre máquinas pode levar a avanços significativos em diversas áreas. Pesquisas recentes têm explorado como a inteligência artificial e o aprendizado de máquina podem ser integrados nesses sistemas, permitindo que os robôs aprendam uns com os outros e adaptem seus comportamentos de maneira autônoma (Cruz, 2023). Essa colaboração entre robôs também facilita a integração de tecnologias emergentes, promovendo maior adaptabilidade e flexibilidade na execução de tarefas complexas, essenciais em ambientes dinâmicos e em constante evolução (Bomfim e Noriega, 2023).

No contexto educacional, por exemplo, os robôs colaborativos podem criar ambientes de ensino e aprendizagem mais interativos e envolventes, facilitando a compreensão de conceitos complexos de maneira acessível. Nesse sentido, os robôs sociais, como NAO e Pepper, têm mostrado ser ferramentas importantes na interação humano-robô, promovendo não apenas o aprendizado técnico, mas também o desenvolvimento de habilidades sociais e emocionais em contextos cotidianos (Sodré, 2023). A capacidade desses robôs de interagir de forma autônoma e natural demonstra seu potencial para contribuir de maneira significativa em várias áreas, especialmente na educação.

Dessa forma, o presente artigo foi organizado em: Referencial teórico, Metodologia, Resultados e Conclusões, O objetivo deste artigo é analisar as interações básicas entre os robôs NAO e Pepper, demonstrando como essas interações podem ser aplicadas para enriquecer o ambiente educacional e social, promovendo a colaboração e o aprendizado.

## 2. TRABALHOS RELACIONADOS

Mubin et al. (2018) em sua pesquisa apresenta uma revisão abrangente sobre o uso de robôs sociais em espaços públicos. Este trabalho analisa como as tecnologias de robótica social evoluíram ao longo do tempo e os desafios que enfrentam em termos de aceitação e eficácia. O estudo enfatiza a importância da interação humano-robô em contextos sociais, destacando que a integração eficaz dos robôs na vida cotidiana depende de vários fatores, como o design, a funcionalidade e a percepção social dos usuários. Os autores ressaltam que, para que os robôs sociais sejam aceitos e utilizados de maneira eficaz, é necessário que sejam projetados considerando não apenas as capacidades técnicas, mas também as expectativas e normas sociais dos ambientes em que operam.

Manzi et al. (2021) investiga as expectativas, atitudes e atribuições mentais de adultos jovens em relação a dois robôs sociais humanoides diferentes. O estudo destaca como esses jovens diferenciam entre os robôs com base em características como *design*, comportamento e funcionalidades, e como essas percepções influenciam a forma como interagem com cada robô. Os resultados mostram que, embora ambos os robôs sejam humanoides, as expectativas e a receptividade variam significativamente, o que sugere que a personalização e o contexto de uso são cruciais para a aceitação e eficácia dos robôs sociais em ambientes educacionais e sociais.

Weike et al. (2024) explora como uma ferramenta de programação visual intuitiva, como o Node-RED, pode ser utilizada para permitir que usuários sem treinamento técnico moldem o comportamento de robôs em cenários de interação humano-robô. O estudo destaca a acessibilidade dessa ferramenta, que possibilita a criação de interações robóticas complexas de maneira intuitiva, tornando a programação robótica mais acessível a um público mais amplo. Os resultados mostram que, ao simplificar o processo de programação, é possível democratizar o desenvolvimento de soluções robóticas em diversos campos, desde a educação até a automação industrial.

Com base nos estudos apresentados, é possível observar como a interação autônoma entre robôs foi aplicada em distintos contextos. Além disso, foi enfatizada a importância de fatores sociais e técnicos que poderiam ser construídos com esses dispositivos. Mubin et al. (2018) destacaram questões relacionadas ao design, interface e funcionalidade que influenciaram na aceitação dos robôs em contextos sociais, essenciais para aplicações educacionais. Enquanto Manzi et al. (2021) mostraram que as percepções dos usuários variavam conforme as capacidades dos robôs, sugerindo a necessidade de adaptar as interações para diferentes públicos. Além disso, Cruz et al. (2023) investigou como a inteligência artificial permitiu que robôs aprendessem uns com os outros, aprimorando suas interações de forma autônoma. Esses trabalhos forneceram uma base relevante para o desenvolvimento de sistemas colaborativos, como NAO e Pepper, associados ao contexto educacional.

## 3. FUNDAMENTAÇÃO TEÓRICA

### 3.1. Robot NAO

O robô **NAO V6**, da **SoftBank Robotics**, foi um dos principais elementos neste estudo. O NAO era conhecido por sua capacidade de interagir de forma natural com humanos, o que ocorria devido aos sensores que incluíam câmeras, microfones e sensores de toque. Além disso, sua aparência antropomórfica facilitava e promovia uma interação mais próxima e intuitiva com os seres humanos. Com relação a aspectos técnicos, este dispositivo apresentava 25 graus de liberdade, o que permite realizar movimentos variados, como andar, levantar os braços e fazer gestos simples parecidos aos realizados por seres humanos. O NAO pode ser visto na Figura 1, e suas especificações detalhadas podem ser consultadas na Tabela 1.

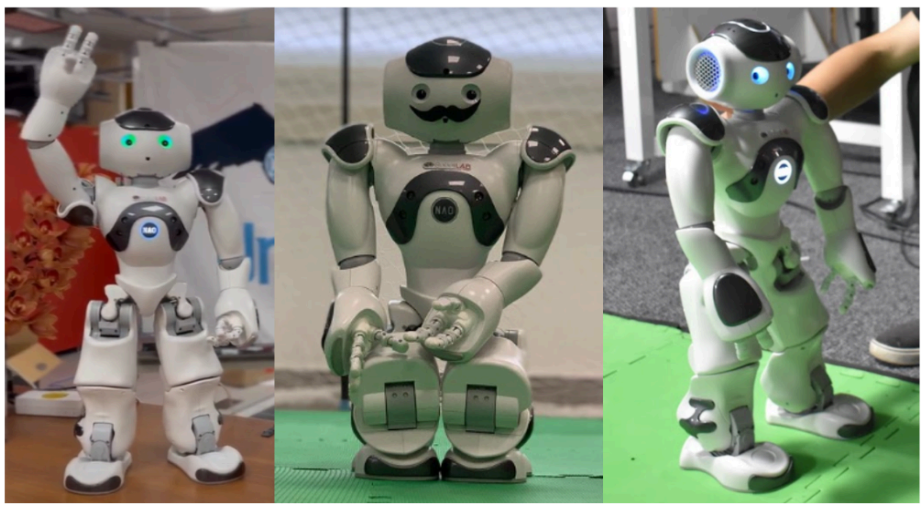

Figura 1. Robot NAO V6.

**Tabela 1. Especificações Robot NAO V6**

| Componentes | Especificações |
|---|---|
| Processador | *Intel ATOM x7-Z8750 1.6 GHz, CPU x86 (SoftBank Robotics Group Corp., 2024)* |
| Câmeras | 2 câmeras HD (1280x720), uma no topo da cabeça e outra na altura da boca. A |

|  | câmera superior é usada para visão de longo alcance e a inferior para visão de curto alcance e reconhecimento facial *(NAO Documentation — Aldebaran 2.1.4.13 Documentation, s. f.).* |
|---|---|
| Sensores | - 4 microfones direcionais para reconhecimento de fala.<br>- 2 alto-falantes para síntesis de fala.<br>- Sensor de pressão em 3 regiões: cabeça, mãos e pés.<br>- Acelerômetros e giroscópios para equilíbrio e orientação.*(NAO Documentation — Aldebaran 2.1.4.13 Documentation, s. f.).* |
| Juntas e Motores | 25 graus de liberdade (DoF) controlados por atuadores, que permitem movimentos complexos (*SoftBank Robotics Group Corp.*, 2024). |
| Formas de Conexão | 25 graus de liberdade (DoF) controlados por atuadores, que permitem movimentos complexos (*SoftBank Robotics Group Corp.*, 2024) |
| Linguagens de Programação | *Wi-Fi, Ethernet, USB 2.0, Bluetooth (NAO Documentation — Aldebaran 2.1.4.13 Documentation, s. f.).* |
| Outros Componentes | *- LEDs RGB nas áreas dos olhos, orelhas e peito.*<br>*- Bateria recarregável de íons de lítio (48.6 Wh) (NAO Documentation — Aldebaran 2.5.11.14a Documentation, s. f.).* |

## 3.2. Robô PEPPER

O robô **Pepper**, também desenvolvido pela **SoftBank Robotics**, foi outro elemento central neste estudo. O robô foi projetado para interações sociais, sendo amplamente utilizado em ambientes públicos, como lojas e eventos, onde a comunicação com humanos é fundamental. O robô Pepper pode reconhecer e responder a emoções humanas utilizando uma combinação de sensores e algoritmos avançados. Equipado com câmeras que capturam expressões faciais e microfones que detectam o tom de voz, o robô processa essas informações através de um software de reconhecimento de emoções, permitindo-lhe identificar estados emocionais como alegria, tristeza ou surpresa. Em resposta, o robô ajusta sua fala, gestos e comportamentos para interagir de maneira mais natural e empática com as pessoas (Amabili et al., 2023). O robô Pepper pode ser visto na Figura 2 e suas especificações detalhadas podem ser consultadas na Tabela 2.

**Tabela 2. Especificaciones Robot Pepper**

| Componentes | Especificações |
|---|---|
| Processador | Intel Atom E3845 1.91 GHz, CPU x86 (Pepper - Documentation — Aldebaran 2.5.11.14a Documentation, s. f.). |
| Câmeras | 2 câmeras HD (1280x720), uma para visão 2D (localizada na cabeça) e outra 3D (localizada na boca), usadas para reconhecimento facial e navegação (Pepper - Documentation — Aldebaran 2.5.11.14a Documentation, s. f.). |
| Sensores | - 4 microfones para captação de som direcional.<br>- 2 alto-falantes para síntese de voz.<br>- 3 sensores de toque (cabeça e mãos).<br>- Giroscópio e acelerômetro para equilíbrio e orientação (Pepper - Documentation — Aldebaran 2.5.11.14a Documentation, s. f.). |
| Juntas e Motores | 20 graus de liberdade (DoF) controlados por atuadores, permitindo movimentos suaves e expressivos (Pepper - Documentation — Aldebaran 2.5.11.14a Documentation, s. f.). |
| Formas de Conexão | Wi-Fi, Ethernet, Bluetooth 4.0, USB 2.0 (Pepper - Documentation — Aldebaran 2.5.11.14a Documentation, s. f.). |
| Linguagens de Programação | Python, C++, Java, Choregraphe (ambiente de programação gráfica), NAOqi SDK (Pepper - Documentation — Aldebaran 2.5.11.14a Documentation, s. f.). |
| Outros Componentes | - LEDs RGB em torno dos olhos para expressões visuais.<br>- Bateria recarregável de íons de lítio com autonomia de até 12 horas.<br>- Sistema de navegação LIDAR integrado para movimentação autônoma |

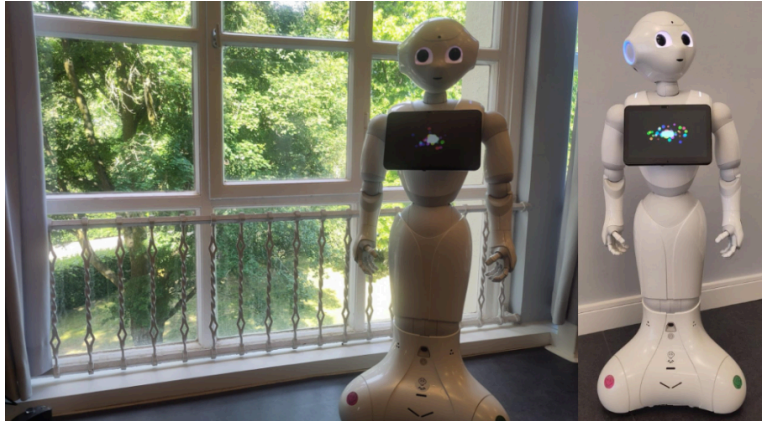

Figura 2. Robot PEPPER.

**3.3. Node-RED**

Para configurar e controlar as interações entre os robôs NAO e Pepper, foi utilizada a ferramenta de programação visual Node-RED[1]. Esta plataforma permite criar e ajustar o comportamento dos robôs através de fluxos de trabalho visuais, que são organizados em blocos de comandos simples e intuitivos (Weike, Ruske, Gerndt, & Doernbach, 2024). A programação das interações foi realizada conectando diferentes nodos que representam as ações e respostas de cada robô Figura 3.

**Node-(RED)²**, uma extensão do Node-RED, foi utilizado para facilitar a integração dos sensores e dispositivos IoT que os robôs utilizam para detectar e responder ao ambiente e entre si. A modularidade da ferramenta permitiu a configuração e o ajuste em tempo real dos comportamentos dos robôs durante os experimentos. Isso garantiu que as interações planejadas fossem executadas de acordo com o esperado, sem a necessidade de programação complexa.

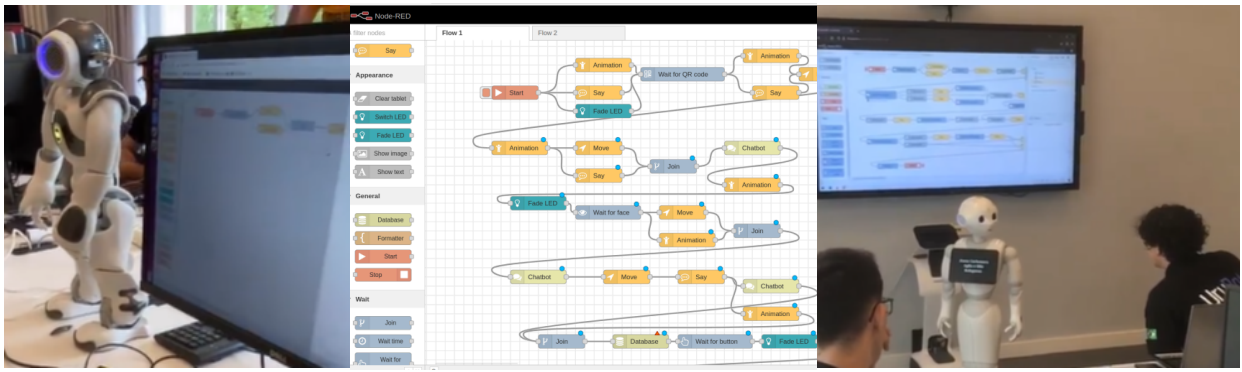

Figura 3. Configurando a interação entre NAO e Pepper utilizando Node-RED.

---

[1] Node-RED é uma ferramenta de programação de fluxo desenvolvida originalmente pela IBM para conectar dispositivos de hardware, APIs e serviços online de maneira modular e visual. É amplamente utilizada em ambientes de Internet das Coisas (IoT) e automação, permitindo a criação de aplicações complexas com uma interface de arrastar e soltar. Para mais informações, visite: Node-RED.

## 4. METODOLOGÍA

Esse trabalho apresentou a construção de um processo de interação de robôs NAO e Pepper dirigido para 14 estudantes de graduação do Uruguai. Esta simulação teve como objetivo principal investigar e aprender sobre as dinâmicas de comunicação entre robôs humanoides em um ambiente educativo e colaborativo.

Os robôs foram programados utilizando a plataforma Node-RED, uma ferramenta de programação visual que facilita a criação de interações complexas de maneira intuitiva (Weike, 2024). As interações permitiram observar como os robôs respondiam e coordenavam suas ações de forma autônoma.

A interação começava com o robô NAO dizendo "*Hello Pepper, how are you*?" enquanto acenava para o robô Pepper. Esse gesto de aceno servia como uma introdução visual e verbal à conversa, simulando o início de uma interação humana. Após essa saudação, os robôs continuavam a troca de frases simples, como saudações e perguntas sobre a localização, imitando uma conversa básica entre dois seres humanos. Durante o diálogo, ambos os robôs realizavam movimentos sincronizados de cabeça e braços, que imitavam gestos humanos, tornando a interação mais natural e envolvente.

A análise dessas interações permitiu identificar aspectos importantes que podem contribuir para o aprimoramento do processo de colaboração e interação entre robôs humanoides. Os estudantes de graduação participaram ativamente, observando como os robôs se comportavam e fornecendo *feedback* sobre a eficácia das interações, o que auxiliou no desenvolvimento de futuras simulações e aplicações práticas no contexto educacional.

## 3. RESULTADOS E DISCUSSÃO

### 3.1. Participantes Colaboradores

O estudo foi realizado na Universidade de Ostfalia, na Alemanha, como parte de um programa de intercâmbio acadêmico que envolveu a equipe de competições em robótica UruBots, composto por 14 estudantes e 2 docentes do Uruguai. O principal objetivo dessa visita à universidade era aprender sobre interação humano-robô, uma área de grande relevância na robótica. Durante o projeto, todos os participantes tiveram a oportunidade de testar códigos e experimentar diferentes abordagens com os robôs NAO e Pepper, explorando como essas interações poderiam ser desenvolvidas e refinadas. A colaboração em grupo foi essencial para o desenvolvimento das interações, e como resultado desse trabalho em conjunto, surgiu a ideia de programar o robô NAO para expressar que estava "perdido", enquanto Pepper respondia indicando que estavam na Alemanha, especificamente na Universidade de Ostfalia Figura 4.

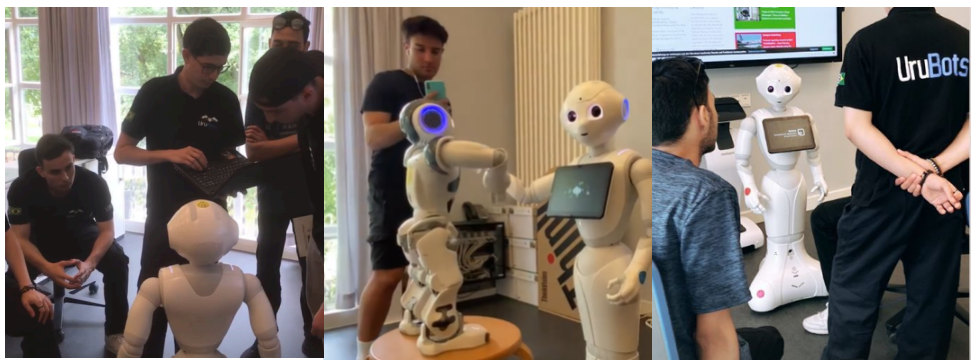

Figura 4. Colaboradores trabalhando com robôs.

### 3.2. Desenvolvimento dos Cenários

O experimento iniciou com a criação de um cenário de Diálogo Contínuo para avaliar a interação verbal entre NAO e Pepper. Durante a preparação deste cenário, foi necessário adaptar o ambiente para garantir que ambos os robôs pudessem se comunicar de maneira eficaz. Como parte dessas adaptações, o robô NAO foi colocado sobre um taburete para que ficasse à altura de Pepper, facilitando a visualização e a captação de áudio entre os dois robôs.

Esta modificação foi importante para melhorar a interação, permitindo que NAO e Pepper se vissem e se escutassem claramente durante o experimento, como pode ser visto na Figura 5.

**Cenário de Diálogo Contínuo:**

- **Objetivo:** Avaliar a capacidade dos robôs de manter um diálogo contínuo e coerente.
- **Descrição:** Neste cenário, NAO e Pepper trocaram frases simples em uma conversa simulada. A colocação de NAO à altura de Pepper facilitou a troca verbal, permitindo que ambos os robôs mantivessem a coerência e a continuidade na interação verbal e nos gestos. Este cenário ajudou a entender como as adaptações físicas no ambiente podem influenciar a qualidade da interação entre robôs humanoides.

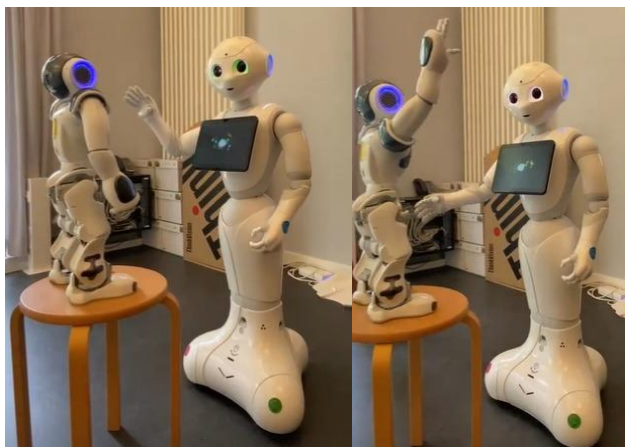

Figura 5. Cenario com NAO e PEPPER.

### 3.3. Interação Verbal entre NAO e Pepper

Durante o experimento, uma das interações programadas envolveu um diálogo simples entre os robôs NAO e Pepper. A conversa consistiu em NAO expressando que estava "perdido" e perguntando onde estavam, ao que Pepper respondeu indicando que estavam na Alemanha, especificamente na Universidade de Ostfalia. Esta interação foi realizada com sucesso, com ambos os robôs trocando frases de maneira fluida e natural. Além de realizar saudações e movimentos coordenados, o NAO também foi programado para "conversar" com o Pepper. Essa interação verbal envolveu uma troca básica de frases, permitindo que os robôs simulassem uma breve conversa. A interação foi contínua, mostrando que os robôs conseguiam manter uma comunicação entre si de forma consistente.

A colocação de NAO em um tamborete permitiu que ambos ficassem na mesma altura, o que facilitou a comunicação visual e auditiva entre os robôs. A seguir está um exemplo da interação:

- **NAO:** *"Hello Pepper, how are you?"*
- **Pepper:** *"Hello NAO, I am fine, how are you?"*
- **NAO:** *"I am lost, where are we?"*
- **Pepper:** *"We are in Germany, specifically at Ostfalia University. Welcome!"*

Além do diálogo verbal, ambos os robôs realizaram gestos sincronizados, como movimentos de cabeça e braços, para simular uma conversa mais humana. Isso contribuiu para uma interação mais rica e envolvente, destacando o potencial dos robôs humanoides em contextos sociais e educativos como pode ser visto na Figura 6.

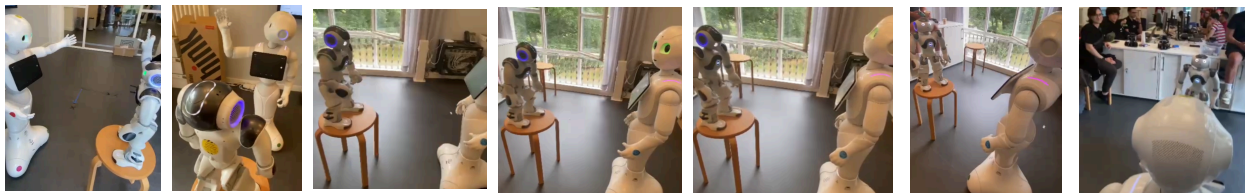

Figura 6. Ciclo de Interação NAO e PEPPER.

Os resultados mostraram que a interação programada entre NAO e Pepper foi bem-sucedida, com ambos os robôs respondendo de maneira adequada e coerente às entradas programadas. A adaptação do ambiente, como a colocação de NAO no taburete, foi fundamental para garantir a qualidade da interação.

**Considerações Éticas:** Durante todo o experimento, foram tomadas medidas para garantir que as interações entre os robôs fossem seguras e respeitassem os princípios éticos, especialmente considerando o uso de robôs em ambientes educacionais.

**Limitações:** Apesar do sucesso do experimento, algumas limitações foram observadas. A necessidade de adaptar o ambiente físico pode não refletir completamente as condições de interações em cenários mais naturais. Além disso, a simplicidade do diálogo programado pode não capturar a complexidade de interações em outros contextos.

## 4. CONCLUSÃO

Este estudo explorou as interações entre os robôs humanoides NAO e Pepper, demonstrando que, com adaptações simples, é possível criar cenários educativos eficazes que promovam tanto o aprendizado técnico quanto o desenvolvimento de habilidades sociais. As experiências realizadas indicaram que esses robôs podem ser integrados com sucesso em ambientes educacionais para apoiar a interação e o desenvolvimento de competências importantes em estudantes, como comunicação, colaboração e resolução de problemas.

Os resultados obtidos sugerem um potencial significativo para expandir e melhorar as interações entre robôs humanoides, especialmente no contexto educacional. Com base nas interações bem-sucedidas observadas durante o experimento, há planos para utilizar esses robôs de maneira mais ampla para ajudar crianças a adquirir habilidades sociais importantes, como comunicação verbal, expressão emocional e cooperação. Além disso, explorar o uso de outros modelos de robôs pode diversificar as oportunidades de aprendizado e adaptar as interações a diferentes necessidades educacionais.

A capacidade de NAO e Pepper de simular interações humanas de maneira natural e envolvente sugere que eles podem ser ferramentas valiosas em ambientes educativos. Assim, futuros estudos devem focar no desenvolvimento de cenários de interação mais complexos e na investigação do impacto dessas tecnologias na educação a longo prazo. A pesquisa indica que o contínuo desenvolvimento e refinamento dessas interações é essencial para maximizar o potencial dos robôs humanoides como ferramentas educativas, contribuindo para a inovação pedagógica e para o preparo dos alunos para os desafios do futuro.

## 6. REFERENCIAS BIBLIOGRÁFICAS